\documentclass[numbers,sort&compress,square]{article}
\usepackage[preprint]{neurips_2025}      
\newcommand{\btheta}{\boldsymbol{\theta}}

\usepackage[]{inputenc}
\usepackage{scalerel}
\usepackage{xcolor}
\usepackage{colortbl}  
\usepackage{placeins}
\usepackage{amsmath, amssymb, amsthm}
\usepackage{booktabs} 
\usepackage{algorithm,algorithmic}
\usepackage{multirow}
\usepackage{graphicx}
\usepackage{float}
\usepackage{wrapfig}
\usepackage{caption}
\usepackage{subcaption}
\usepackage{enumitem}
\usepackage[colorlinks=true, linkcolor=blue, citecolor=blue, urlcolor=blue]{hyperref}

\title{Avoid Forgetting by Preserving Global Knowledge Gradients in Federated Learning with Non-IID Data}

\author{Abhijit Chunduru$^{*1}$, Majid Morafah$^{*2}$, Mahdi Morafah$^{3}$, Vishnu Pandi Chellapandi$^{4}$, Ang Li$^{5}$
\\\\
$^{1}$University of Massachusetts Amherst, $^{2}$Arizona State University, \\
$^{3}$University of California San Diego, $^{4}$Purdue University, $^{5}$University of Maryland College Park\\
$^{*}$Equal first authorship and contribution
}

\begin{document}

\maketitle

\begin{abstract}
The inevitable presence of data heterogeneity has made federated learning very challenging. There are numerous methods to deal with this issue, such as local regularization, better model fusion techniques, and data sharing. Though effective, they lack a deep understanding of how data heterogeneity can affect the global decision boundary. In this paper, we bridge this gap by performing an experimental
analysis of the learned decision boundary using a toy example. Our observations are surprising: (1) we find that the existing methods suffer from forgetting and clients forget the global decision boundary and only learn the perfect local one, and (2) this happens regardless of the initial weights, and clients forget the global decision boundary even starting from pre-trained optimal weights. In this paper, we present FedProj, a federated learning framework that robustly learns the global decision boundary and avoids its forgetting during local training. To achieve better ensemble knowledge fusion, we design a novel server-side ensemble knowledge transfer loss to further calibrate the learned global decision boundary. To alleviate the issue of learned global decision boundary forgetting, we further propose leveraging an episodic memory of average ensemble logits on a public unlabeled dataset to regulate the gradient updates at each step of local training. Experimental results demonstrate
that FedProj outperforms state-of-the-art methods by a large margin.
\end{abstract}

\section{Introduction}

Federated Learning (FL) \cite{kairouz2021advances} has emerged as a privacy-preserving framework that trains a shared global model across multiple clients without exchanging raw data, under the coordination of a central server. The most widely adopted FL method, FedAvg \cite{fedavg}, aggregates local weight or gradient updates to learn the global model. Although FedAvg has demonstrated promise across various applications, its performance is significantly hindered by data heterogeneity \cite{glasgow2022sharp,woodworth2020minibatch,li2019convergence,chellapandi2023federated,chellapandi2023convergence,chellapandi2023survey,chellapandi2024fednmut,morafah2024stable,morafah2025clustered,morafah2024large} among clients—commonly referred to as Non-IID data—which can lead to poor convergence and degraded performance.

Numerous studies have sought to address the challenges posed by Non-IID data in FL by mitigating client drift \cite{fedprox,karimireddy2020scaffold}, enhancing server-side model fusion and distillation \cite{cheng2021fedgems,zhu2021data,li2019fedmd}, and refining local training protocols \cite{collins2021exploiting}. Despite these efforts, a critical issue remains underexplored: during local training, clients tend to overfit to their individual objectives, thereby \emph{catastrophically forgetting the global knowledge and decision boundaries learned by the shared aggregated model}. Although prior works have acknowledged this phenomenon \cite{yang2024federated}, a comprehensive experimental investigation quantifying global knowledge forgetting is lacking. In our work, we first present an empirical pilot study of standard federated learning, i.e. FedAvg, under Non-IID settings, demonstrating that local clients often completely forget global knowledge and converge to models that are fine-tuned solely to their local objectives. Consequently, \emph{averaging such divergent models merely restores the original performance rather than enhancing it}, and the server-side fusion is further compromised as client models have become increasingly dissimilar. Notably, this catastrophic forgetting occurs even when clients begin training from a pre-trained global model.

Motivated by these insights we propose a novel federated learning method called \emph{FedProj}. Our approach mitigates global knowledge forgetting by imposing explicit constraints on local gradient updates where the actual forgetting happens, thereby preserving the learned global knowledge. Specifically, we maintain a small episodic memory of global knowledge using a public dataset and formulate an optimization problem that constrains new gradient directions to prevent an increase in the losses associated with the global knowledge. At the server side, we further enhance model fusion through ensemble knowledge distillation using both logits and feature representations, and we incorporate a novel weight divergence regularization term to alleviate the adverse effects of noisy distillation. Extensive experiments on both computer vision and natural language processing tasks demonstrate that FedProj significantly outperforms state-of-the-art methods across multiple datasets, achieving superior performance in Non-IID federated learning scenarios.

\textbf{Contributions.} Our work makes the following contributions: 

\begin{itemize}[leftmargin=*]
\item We present the first empirical pilot study that provides new insights into the phenomenon of catastrophic forgetting in standard federated learning under data heterogeneity. 

\item We introduce a novel federated learning method, \emph{FedProj}, which mitigates global knowledge forgetting by imposing explicit constraints on local gradient updates and preserving global knowledge.

\item We conduct extensive experiments on both CV and NLP tasks and demonstrate that our proposed method outperforms state-of-the-art approaches on various benchmarks and datasets under Non-IID settings.
\end{itemize}

\textbf{Organization.} The remainder of this paper is organized as follows. In Section 2, we review the related work. Section 3 presents our empirical pilot study. In Section 4, we describe the proposed {FedProj} methodology. Section 5 details our experimental results, and Section 6 concludes the paper. 


\section{Related Works}

\textbf{Continual Learning.}
In continual learning, gradient projection methods mitigate catastrophic forgetting by constraining parameter updates to preserve past knowledge. These approaches identify critical directions in weight space and restrict updates to be orthogonal or complementary to them \cite{OWM,OGD}. In particular, Gradient Projection Memory (GPM) \cite{GPM} focuses on key gradient subspaces and explicitly preserves the past gradient subspaces. More recent methods relax strict orthogonality to balance stability and plasticity \cite{TRGP, SGP}. These methods frame continual learning as a stability-plasticity tradeoff in linear algebra, ensuring knowledge retention while allowing adaptability. Our work draws intuition from these body of works in continual learning to allieviate the catastrophic forgetting in FL.


\noindent\textbf{FL with Non-IID Data.} 
Federated learning with non-IID data presents significant challenges, prompting diverse strategies to improve convergence and accuracy. FedAvg \cite{fedavg} established the foundation for FL by averaging local model updates, but it struggles with statistical heterogeneity. To address this, FedProx \cite{fedprox} introduces a proximal term to stabilize local updates, while FedNova \cite{fednova} normalizes local contributions, mitigating objective inconsistency. SCAFFOLD \cite{karimireddy2020scaffold} introduces control variates to correct local updates, effectively reducing drift by anchoring clients to a global direction.  FedGen \cite{fedgen} tackles non-IID challenges by generating synthetic data at the server using generative models, improving generalization across clients. For personalization, Ditto \cite{li2021ditto} maintains dual local-global models, improving client-specific performance, while pFedHN \cite{pfedhn} uses hypernetworks to generate personalized models, enhancing adaptability. FedBN \cite{fedbn} addresses feature shift by keeping batch normalization layers local, while sharing other parameters, boosting robustness in non-IID settings. MOON \cite{moon} employs contrastive learning to align local and global models, reducing divergence.

\textbf{Comparison with Close Works.}
FedDF \cite{feddf} enhances server-side performance through knowledge distillation, refining the global model by distilling knowledge from local models. FedET \cite{fedet} employs ensemble distillation, combining local knowledge into a generalized global model, improving accuracy on heterogeneous data. FedGKT \cite{fedgkt} introduces group knowledge transfer, where lightweight client models distill knowledge into a larger server model, boosting efficiency and handling model heterogeneity. 
Researchers in \cite{gong2022preserving} propose a privacy-preserving FL framework with one-shot offline knowledge distillation using unlabeled public data, reducing communication overhead while enhancing privacy guarantees.

In contrast, FedProj directly applies gradient projection onto the global knowledge gradient subspace, effectively mitigating client drift and preserving global knowledge across non-IID clients. 
Moreover, while FedDF, FedET, and FedGKT focus on logit-level or ensemble-based distillation, FedProj retains knowledge at the gradient level, ensuring more effective knowledge alignment in non-IID FL settings.
\begin{figure*}[ht]
    \centering
     \includegraphics[width=.99\linewidth]{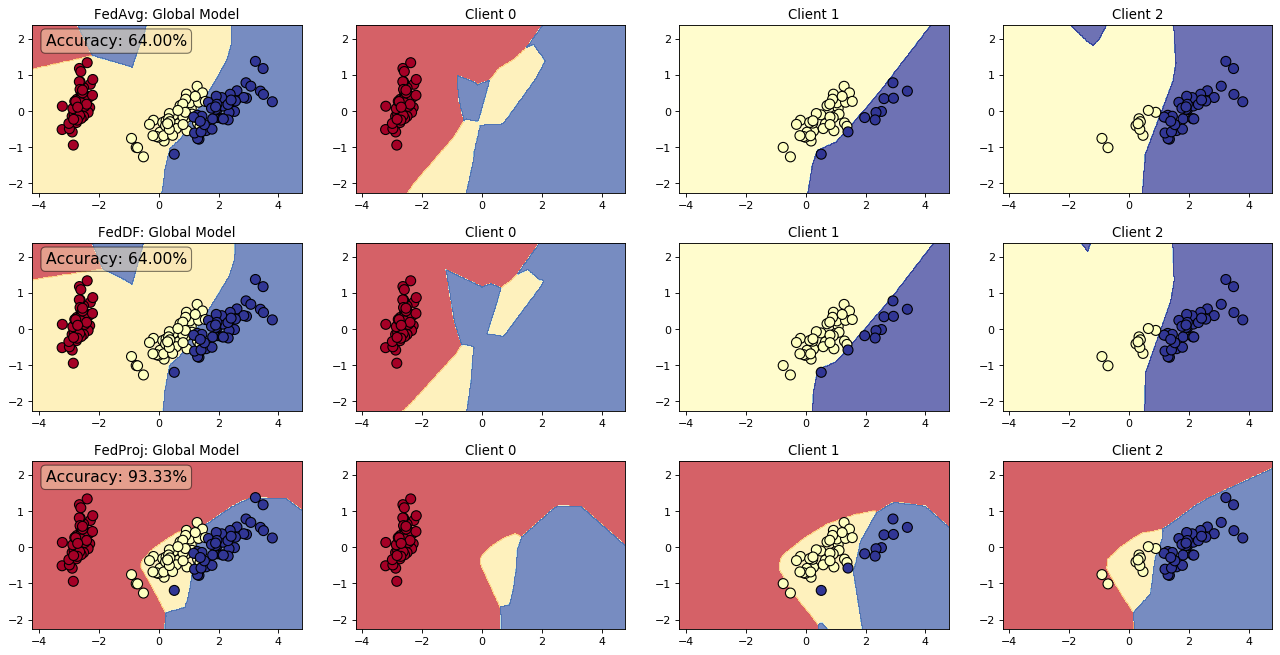}

    \caption{\textbf{Visualization of Catastrophic Forgetting of Global Decision Boundaries under Non-IID Federated Learning.} This figure illustrates that standard federated learning methods (FedAvg and FedDF) experience significant catastrophic forgetting of global decision boundaries after local training on Non-IID client data, leading to poor global model performance (64\%). In comparison, our proposed FedProj method successfully preserves global decision boundaries during local updates, resulting in a robust and highly accurate global model (93.33\%).}
    \label{fig1}
    \vspace{-0.8cm}
\end{figure*}

\section{Pilot Study: Catastrophic Forgetting of Global Knowledge}

In this section, we conduct a systematic empirical study to demonstrate and analyze catastrophic forgetting of global knowledge, specifically global decision boundaries, in federated learning under Non-IID data conditions.

\noindent\textbf{Experimental Setup.} To clearly visualize decision boundaries and intuitively illustrate global knowledge forgetting, we utilize the Iris dataset and apply Principal Component Analysis (PCA) to reduce its dimensionality to 2D. The Iris dataset comprises 150 samples evenly distributed among three distinct classes: Setosa, Versicolor, and Virginica, each characterized by four numerical features. We partition the data into three Non-IID clients, each predominantly containing data from different subsets of the available classes to simulate realistic heterogeneous conditions.

We employ a simple 3-layer Multi-Layer Perceptron (MLP) neural network as the local model at each client. For optimization, we use Stochastic Gradient Descent (SGD) with a learning rate of $1\times10^{-3}$ and momentum of 0.9. To compare standard FL methods with our proposed method, we run federated learning using FedAvg, FedDF, and FedProj. We set the number of communication rounds to 20 and allow each client to perform local training for 5 epochs per round.

\noindent\textbf{Results and Analysis.} Figure~\ref{fig1} presents visualizations of decision boundaries generated by each method. The decision boundaries for each class—red, yellow, and blue—are indicated by their corresponding colors. The client-specific decision boundaries are plotted after the completion of local training for each communication round.

Our observations clearly indicate catastrophic forgetting in standard federated learning methods (FedAvg and FedDF). After local training, individual clients drastically diverge from the global decision boundaries, particularly evident in Client 1 and Client 2. This divergence signifies the loss of global knowledge as clients overfit to their localized data distribution, ignoring global classification objectives. Consequently, aggregating these diverged local models leads to poor global accuracy and unstable decision boundaries.

In contrast, our proposed method, FedProj, demonstrates significant robustness against global knowledge forgetting. By explicitly incorporating constraints via gradient projection to retain global decision boundaries, FedProj effectively preserves global knowledge throughout the local training process. As a result, the aggregated global model consistently maintains high classification accuracy and stable decision boundaries across all clients.

\section{Methodology} \label{sec:method}
In this section, we detail our proposed \emph{FedProj} methodology, a novel federated learning gradient projection algorithm that addresses \emph{global knowledge catastrophic forgetting} in the presence of Non-IID data. Our approach combines explicit {constraint-based gradient projection} in client-side local training with server-side {knowledge distillation}. Below, we first describe the basic problem setting and provide a mathematical formulation of our goal, then detail the proposed method.

\subsection{Problem Formulation and Preliminaries}
\label{subsec:prob}
Consider a federated learning system with $N$ clients, indexed by $k=1,\dots,N$, each with a private local dataset $\mathcal{D}_k$. Let $\btheta \in \mathbb{R}^p$ denote the parameters of the global model to be learned. In standard federated learning, the server initializes $\btheta_g^{(0)}$ and, at each communication round $t$, sends the current global parameters $\btheta_g^{(t)}$ to a subset of selected clients $\mathcal{S}_t \subseteq \{1,\dots,N\}$. Each client $k\in \mathcal{S}_t$ then performs local training using its private data $\mathcal{D}_k$, yielding a locally updated model $\btheta_k^{(t+1)}$. The server then aggregates these updates (e.g., via simple averaging):

\begin{align}
\label{eq:fedavg}
\btheta_g^{(t+1)} \;=\; \sum_{k \in \mathcal{S}_t} \frac{\lvert \mathcal{D}_k \rvert}{\sum_{j \in \mathcal{S}_t}\lvert \mathcal{D}_j \rvert}\,\btheta_k^{(t+1)}.
\end{align}

While such aggregation works well under IID settings, it suffers from severe performance drops under data heterogeneity (Non-IID). Our \emph{FedProj} algorithm addresses this by incorporating:

\begin{enumerate}
    \item \textbf{Client-Side Gradient Projection:} During local updates, we impose explicit gradient constraints that preserve \emph{global knowledge}, thereby preventing the local model from overfitting exclusively to client-specific objectives.
    \item \textbf{Server-Side Knowledge Distillation:} Once local updates are sent back, the server fuses them more effectively through ensemble distillation, using a small, auxiliary \emph{public dataset} to align their logits and feature representations. Furthermore, to avoid the negative impact of noisy ensemble distillation we further add a new weight divergence regularizer.
\end{enumerate}

\subsection{Local Training with Gradient Projection}
\label{subsec:client}
To illustrate the core idea of our gradient projection, we first define the local objective and then show how to preserve global knowledge via an explicit constraint on the gradient.

\textbf{Local Objective.} Let $k$ be a particular client in the selected subset $\mathcal{S}_t$ at round $t$. The client receives $\btheta_g^{(t)}$ from the server and initializes $\btheta_k$ with it. The local training objective typically involves empirical risk minimization on $\mathcal{D}_k$, for instance:
\begin{equation}
\label{eq:local_obj}
\min_{\btheta_k} \quad \frac{1}{\lvert \mathcal{D}_k \rvert} \sum_{(\mathbf{x},\mathbf{y}) \in \mathcal{D}_k} \ell\bigl(f(\mathbf{x};\,\btheta_k),\;\mathbf{y}\bigr),
\end{equation}
where $\ell(\cdot,\cdot)$ is a loss function (e.g., cross-entropy) and $f(\mathbf{x};\,\btheta_k)$ is the client-side model output.

\noindent\textbf{Global Knowledge Memory.}
Recall that our objective is to avoid \emph{global knowledge forgetting} during each client's local updates. To achieve this, we introduce a \emph{memory loss} \(\mathcal{L}_{\mathrm{mem}}\) that measures how well the current local model \(\btheta_k\) preserves the global knowledge. Specifically, the global knowledge is distilled via the server’s ensemble logits on a small public dataset \(\mathcal{D}_{\mathrm{pub}}\). Let
\[
\mathcal{Z}_{\mathrm{server}}(\mathbf{x}_m) 
  \;=\; \frac{1}{|\mathcal{S}_t|}\,\sum_{\ell\in \mathcal{S}_t}\,f_{\ell}(\mathbf{x}_m;\,\btheta_\ell^{(t+1)}),
\]
be the server-aggregated logits (averaged over the selected clients \(\mathcal{S}_t\)) for a sample \(\mathbf{x}_m\in \mathcal{D}_{\mathrm{pub}}\). We denote the local model logits by \(\mathcal{Z}_{k}(\mathbf{x}_m;\,\btheta_k)\). Then we define 

\begin{equation}
  \label{eq:local_mem_loss}
  \begin{split}
    \mathcal{L}_{\mathrm{mem}}(\btheta_k)
    &= \frac{1}{|\mathcal{M}|} \sum_{(\mathbf{x}_m)\in\mathcal{M}}
    \mathrm{KL}\bigl(\sigma(\mathcal{Z}_{\mathrm{server}}(\mathbf{x}_m)), \\
    &\quad\sigma(\mathcal{Z}_{k}(\mathbf{x}_m;\,\btheta_k))\bigr)
  \end{split}
\end{equation}

where \(\mathcal{M}\subseteq \mathcal{D}_{\mathrm{pub}}\) is a small \emph{memory buffer} for preserving global knowledge, \(\sigma(\cdot)\) is the softmax function, and \(\mathrm{KL}\) is the Kullback--Leibler divergence. Minimizing $\mathcal{L}_{\mathrm{mem}}$ ensures that local updates remain aligned with global model knowledge, preventing catastrophic forgetting.

\noindent\textbf{Constrained Local Objective and Gradient Projection.}
Let \(\mathcal{L}_{\mathrm{local}}(\btheta_k)\) be the ordinary local objective (e.g., cross-entropy on the client’s private data \(\mathcal{D}_k\)). Our goal is to minimize \(\mathcal{L}_{\mathrm{local}}\) \emph{without} increasing \(\mathcal{L}_{\mathrm{mem}}\). Formally, we can write a constraint-based objective:
\begin{align} 
\min_{\btheta_k}\;\;\mathcal{L}_{\mathrm{local}}(\btheta_k)
\quad
\text{subject to}
\quad
\mathcal{L}_{\mathrm{mem}}(\btheta_k)\;\le\;\mathcal{L}_{\mathrm{mem}}\!\bigl(\btheta_k^{\mathrm{old}}\bigr),
\label{eq:constraint_obj}
\end{align}
where \(\btheta_k^{\mathrm{old}}\) is the local model state prior to the new update (i.e., just received from the server). Intuitively, the local model’s memory loss must not exceed its old memory loss, ensuring that the updated local parameters \(\btheta_k\) do not degrade global knowledge.



Although \eqref{eq:constraint_obj} is straightforward conceptually, it is challenging to solve directly in the high-dimensional parameter space of neural networks. Instead, we locally approximate \(\mathcal{L}_{\mathrm{mem}}(\btheta_k)\) around \(\btheta_k^{\mathrm{old}}\). Concretely, let:
\begin{align*}
g_{\text{new}} \;=\;\nabla_{\btheta_k}\,\mathcal{L}_{\mathrm{local}}(\btheta_k), 
\quad
g_{\mathrm{glob}} \;=\;\nabla_{\btheta_k}\,\mathcal{L}_{\mathrm{mem}}(\btheta_k).
\end{align*}
Requiring \(\mathcal{L}_{\mathrm{mem}}(\btheta_k)\le \mathcal{L}_{\mathrm{mem}}(\btheta_k^{\mathrm{old}})\) to first order amounts to ensuring \(\langle g_{\text{proj}},\,g_{\mathrm{glob}}\rangle \ge 0\). Therefore, we rewrite the constraint in terms of the projected gradient update \(g_{\text{proj}}\), leading to the following equivalent quadratic program:
\begin{align}
\label{eq:constraint_final}
    \min_{g_{\text{proj}}} & \quad \frac{1}{2}\bigl\|\,g_{\text{new}} \;-\; g_{\text{proj}}\bigr\|^2_2, \quad
    \text{subject to}\quad \langle g_{\text{proj}},\,g_{\mathrm{glob}}\rangle \;\ge\; 0.
\end{align}
Solving \eqref{eq:constraint_final} enforces that our final update direction does not \emph{negatively correlate} with \(g_{\mathrm{glob}}\), thus preventing increases in the memory-based loss and preserving the global knowledge as a result.

The optimal $g_{\text{proj}}$ can be derived via standard Lagrangian methods, yielding:
\begin{align*}
\label{eq:proj_grad_optimal}
g_{\text{proj}} \;=\;
\begin{cases}
    g_{\text{new}}, &\quad\text{if } \langle g_{\text{new}}, g_{\mathrm{glob}}\rangle \geq 0, \\[5pt]
    g_{\text{new}} - \frac{\langle g_{\text{new}},\,g_{\mathrm{glob}}\rangle}{\|\,g_{\mathrm{glob}}\|^2 + \epsilon}\,g_{\mathrm{glob}}, &\quad\text{otherwise},
\end{cases}
\end{align*}
where \(\epsilon>0\) is added for numerical stability. Finally, the local model parameters are updated as:
\begin{equation}
\label{eq:local_update_final}
\btheta_k \leftarrow \btheta_k \;-\;\eta_{\mathrm{local}}\,g_{\text{proj}},
\end{equation}
where $\eta_{\mathrm{local}}$ is the local learning rate. This ensures that we locally minimize \(\mathcal{L}_{\mathrm{local}}\) while preserving global knowledge encapsulated by \(\mathcal{L}_{\mathrm{mem}}\).


\begin{figure}[ht]
    \centering
     \includegraphics[width=0.8\linewidth]{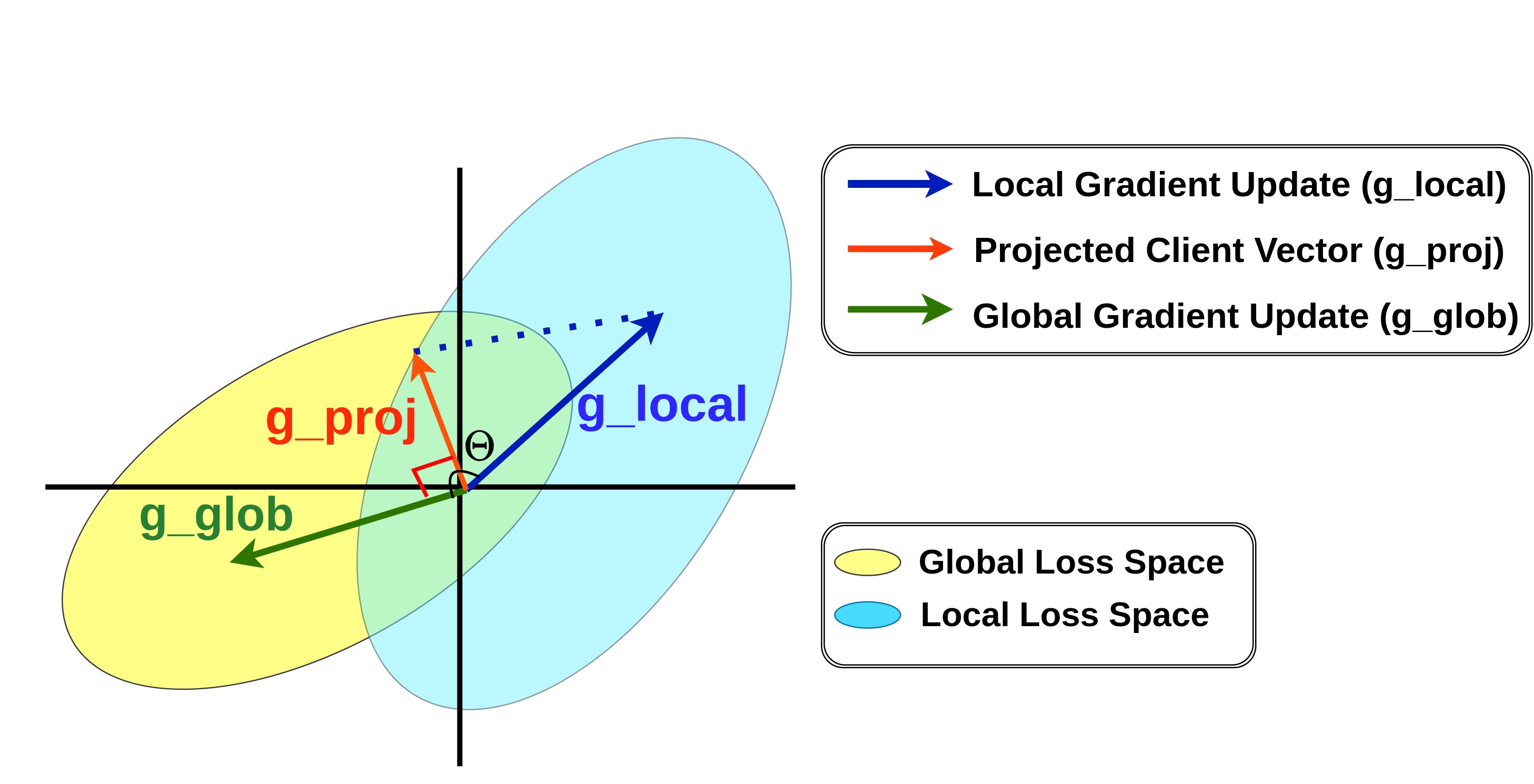}

    \caption{\textbf{Gradient Projection in FedProj:} Local gradient updates ($g_{\text{local}}$) are projected onto a subspace orthogonal to the global gradient ($g_{\text{glob}}$), resulting in the projected vector ($g_{\text{proj}}$) for better knowledge retention.}
    \label{fig2}
\end{figure}

Figure~\ref{fig2} illustrates the gradient update process in heterogeneous local and global loss spaces. The yellow and blue regions represent the global and local loss spaces, respectively. The local gradient update, \(g_{\text{local}}\) (solid blue arrow), may deviate from the global objective due to data heterogeneity. To correct this, it is projected onto the global loss space, yielding \(g_{\text{proj}}\) (solid red arrow), which better aligns with the global objective. The final update, \(g_{\text{glob}}\) (green arrow), integrates the projected gradient, preserving global knowledge. The angle \(\theta\) highlights the deviation, while the red arc indicates the correction applied.

The full details of FedProj algorithm (Algorithm \ref{alg:FedProj}) and mathematical derivations are presented in the Appendix.

\subsection{Server-Side Distillation and Model Fusion}
\label{subsec:server}
After all participating clients finish their local updates, the server aggregates them. Instead of a simple weight average stated in equation~\eqref{eq:fedavg}, {FedProj} further \emph{distills} knowledge across clients via a small \emph{public} or \emph{proxy} dataset. Let $\{\btheta_k^{(t+1)}\}_{k\in \mathcal{S}_t}$ be the client-updated parameters. We form an \emph{ensemble of teachers}
\[
\{\;f_k(\cdot;\btheta_k^{(t+1)}) \;\mid\;k\in\mathcal{S}_t\}\,,
\]
and a \emph{student model} $f_g(\cdot;\,\btheta_g)$ on the server, initially set to the simple FedAvg result $\btheta_g = \btheta_g^{\mathrm{FedAvg}}$ obtained via equation~\eqref{eq:fedavg}.

\textbf{Logit Distillation.} For each mini-batch $\mathbf{X}$ sampled from $\mathcal{D}_{\mathrm{pub}}$, we compute teacher logits:
\[
\mathcal{Z}_{\mathrm{teacher}} \;=\; \frac{1}{|\mathcal{S}_t|}\sum_{k \in \mathcal{S}_t}\;f_k(\mathbf{X};\btheta_k^{(t+1)}),
\]
then align the student’s logits $\mathcal{Z}_{\mathrm{student}} = f_g(\mathbf{X};\btheta_g)$ to these teacher logits through a knowledge-distillation loss:
\begin{align}
\label{eq:kd_loss}
\mathcal{L}_{\mathrm{KD}} \;=\;
\mathrm{KL}\Bigl(\sigma\bigl(\mathcal{Z}_{\mathrm{student}}\bigr)\;,\;\sigma\bigl(\mathcal{Z}_{\mathrm{teacher}}\bigr)\Bigr),
\end{align}
where $\mathrm{KL}(\cdot,\cdot)$ is the Kullback–Leibler divergence, and $\sigma(\cdot)$ is the softmax function.


\textbf{Weight Divergence Regularization.} The ensemble distillation process is noisy. This noise is primarily due to utilizing public data that is different from the actual learning private dataset. To scaffold the global model from the nosiy ensemble distillation process, we introduce a weight divergence regularization term. This term encourages the global model parameters to remain close to the averaged model in order to not forget the learned knowledge during the noisy distillation process. In particular, this regularization term allows for learning new knowledge while not being negatively impacted by the noise. Formally, given the global model parameters $\btheta_g$ and the local model parameters $\btheta_k$, one can impose:
\begin{align*}
    \mathcal{L}_{\mathrm{div}} = \,\bigl\|\btheta_k - \btheta_g\bigr\|_2^2.
\end{align*}
\textbf{Overall Server-Side Objective.} Combining logit distillation, feature distillation, and weight divergence regularization, the server update solves:
\begin{align}
\label{eq:server_loss}
\min_{\btheta_g}\quad
\mathcal{L}_{\mathrm{KD}}(\btheta_g) \;+\; \alpha \mathcal{L}_{\mathrm{div}}(\btheta_g),
\end{align}

using gradient-based optimization for $E_d$ distillation epochs at the server, with learning rate $\eta_{\mathrm{distill}}$. The final server parameters become $\btheta_g^{(t+1)}$.

\section{Experiments} \label{sec:experiments}

\subsection{Main Experimental Setup} 
\noindent\textbf{Datasets and Architecture} \label{subsec:Datasets}
We evaluate our approach across computer vision (CV) and natural language processing (NLP) tasks. For CV, we perform image classification on CIFAR-10/100 \cite{cifar} and CINIC-10 \cite{cinic10}. For NLP, we fine-tune pre-trained models on MNLI \cite{mnli}, SST-2 \cite{sst2} and MARC \cite{marc}. We employ ResNet-8 for CIFAR-10, ResNet-18 for CIFAR-100 and CINIC-10, and Tiny BERT \cite{tinybert} for NLP tasks. Data heterogeneity is simulated via Dirichlet distribution \cite{dirchlet} with concentration parameters $\beta \in \{0.3, 0.5\}$.

\noindent\textbf{Federated Learning Setup.} \label{subsec:fl_setup}
For CV tasks, we use 100 clients for CIFAR-10, and CINIC-10 datasets  and 50 clients for CIFAR-100 dataset, fixing client sampling rate to 10\% for all cases. Training spans 100 rounds for CIFAR datasets and 60 for CINIC-10, and local epoch is fixed to 20 for all cases. NLP experiments involve 15 clients with 30\% client sample rate, 1 local epoch, and 15 communication rounds. For server-side distillation, we utilized auxiliary datasets: CIFAR-100 for CIFAR-10/CINIC-10, ImageNet-100 \cite{imagenet} for CIFAR-100, SNLI \cite{snli} for MNLI, Sentiment140 \cite{sent140} for SST-2, and Yelp \cite{yelp} for MARC.

\noindent\textbf{Implementation Details.} \label{subsec:implementation_details} The code is implemented in PyTorch 2.4.1 and executed on NVIDIA RTX 3090 GPUs, using the FedZoo benchmark~\cite{morafah2023practical}. The implementation is anonymously available at \url{https://anonymous.4open.science/r/FedProj_Neurips-63C1}.
We use Adam optimizer with learning rate 0.001 for CV and $3\times10^{-5}$ for NLP tasks. Server-side distillation employs KL divergence loss with temperature $T=3$, performed for 1 epoch (CIFAR-10, CINIC-10, all NLP) or 3 epochs (CIFAR-100) with batch sizes of 256 (CV) and 128 (NLP).

\noindent\textbf{Baselines and Evaluation.} \label{subsec:baselines}
We compare \textsc{FedProj} against established methods: \textsc{FedAvg}~\cite{fedavg}, \textsc{FedProx}~\cite{fedprox}, \textsc{FedNova}~\cite{fednova}, \textsc{FedDF}~\cite{feddf}, \textsc{FedET}~\cite{fedet}, \textsc{MOON}~\cite{moon}, \textsc{FedDyn}~\cite{feddyn}, and \textsc{FedRCL}~\cite{RCL}. Results report average performance and standard deviation across three independent runs with different random seeds, evaluated using global classification accuracy on held-out test sets.

\subsection{Main Experimental Results}
\noindent\textbf{Performance on CV Task.}
Table~\ref{tab:fl_results} presents performance across datasets under non-i.i.d. conditions ($\beta=0.3$ and $\beta=0.5$). FedProj consistently outperforms all baselines, achieving 65.52\% and 69.88\% on CIFAR-10, 35.27\% and 38.06\% on CIFAR-100, and 41.46\% and 41.63\% on CINIC-10, respectively. While MOON performs well under moderate heterogeneity ($\beta=0.5$), it degrades sharply under higher skew ($\beta=0.3$). FedProj maintains robust performance through its projection-based update rule, which aligns local gradients with global descent direction. On the more complex CIFAR-100, alternatives like FedNova and FedDyn exhibit lower accuracy and higher variance. For CINIC-10, which introduces domain shift, FedProj outperforms distillation-based methods like FedDF and FedET. These results demonstrate that FedProj's integration of global memory, gradient projection, and dual-mode distillation effectively addresses client heterogeneity.

\begin{table*}[ht]
\centering
\caption{\textbf{Performance on CIFAR-10, CIFAR-100, and CINIC-10} under Dirichlet($\beta=0.3$, $0.5$).}
\label{tab:fl_results}
\resizebox{\textwidth}{!}{%
\begin{tabular}{l|cc|cc|cc}
\toprule
\textbf{Baseline} & \multicolumn{2}{c|}{\textbf{CIFAR-10}} & \multicolumn{2}{c|}{\textbf{CIFAR-100}} & \multicolumn{2}{c}{\textbf{CINIC-10}} \\
 & \textbf{Dir($\beta$=0.3)} & \textbf{Dir($\beta$=0.5)} & \textbf{Dir($\beta$=0.3)} & \textbf{Dir($\beta$=0.5)} & \textbf{Dir($\beta$=0.3)} & \textbf{Dir($\beta$=0.5)} \\
\midrule
FedAvg \cite{fedavg}    & 63.19 {\tiny$\pm$ 1.48} & 66.41 {\tiny$\pm$ 0.56} & 33.72 {\tiny$\pm$ 0.17} & 37.18 {\tiny$\pm$ 0.09} & 40.59 {\tiny$\pm$ 0.12} & 40.70 {\tiny$\pm$ 0.18} \\
FedProx \cite{fedprox}    & 61.40 {\tiny$\pm$ 0.92} & 67.34 {\tiny$\pm$ 0.36} & 33.96 {\tiny$\pm$ 0.88} & 36.66 {\tiny$\pm$ 0.49} & 40.69 {\tiny$\pm$ 0.07} & 40.80 {\tiny$\pm$ 0.17} \\
FedNova \cite{fednova}   & 63.43 {\tiny$\pm$ 0.99} & 67.93 {\tiny$\pm$ 0.49} & 33.40 {\tiny$\pm$ 0.55} & 36.40 {\tiny$\pm$ 0.48} & 39.81 {\tiny$\pm$ 0.25} & 40.01 {\tiny$\pm$ 0.18} \\
FedDyn \cite{feddyn}    & 63.35 {\tiny$\pm$ 1.03} & 67.53 {\tiny$\pm$ 0.71} & 33.61 {\tiny$\pm$ 0.36} & 36.52 {\tiny$\pm$ 0.39} & 40.43 {\tiny$\pm$ 0.11} & 40.59 {\tiny$\pm$ 0.10} \\
MOON \cite{moon}    & 61.09 {\tiny$\pm$ 1.36} & 68.83 {\tiny$\pm$ 0.78} & 30.29 {\tiny$\pm$ 0.71} & 34.49 {\tiny$\pm$ 0.09} & 40.64 {\tiny$\pm$ 0.10} & 40.79 {\tiny$\pm$ 0.14} \\
FedRCL \cite{RCL}     & 62.14 {\tiny$\pm$ 0.51 } & 67.26 {\tiny$\pm$ 0.87} & 33.91{\tiny$\pm$ 0.20} &  36.77{\tiny$\pm$ 0.11 } &  40.72{\tiny$\pm$ 0.08} &  40.81{\tiny$\pm$ 0.16} \\
FedDF \cite{feddf}     & 61.32 {\tiny$\pm$ 2.02} & 67.44 {\tiny$\pm$ 0.83} & 33.65 {\tiny$\pm$ 0.65} & 36.60 {\tiny$\pm$ 0.30} & 39.34 {\tiny$\pm$ 0.37} & 39.57 {\tiny$\pm$ 0.11} \\
FedET \cite{fedet}     & 58.79 {\tiny$\pm$ 1.20} & 66.46 {\tiny$\pm$ 0.33} & 32.85 {\tiny$\pm$ 0.31} & 36.21 {\tiny$\pm$ 0.14} & 39.11 {\tiny$\pm$ 0.21} & 39.20 {\tiny$\pm$ 0.12} \\
\cellcolor{cyan!15}\textbf{FedProj} & \cellcolor{cyan!15}\textbf{65.52} {\tiny$\pm$ 0.86} & 
\cellcolor{cyan!15}\textbf{69.88} {\tiny$\pm$ 0.03} & 
\cellcolor{cyan!15}\textbf{35.27} {\tiny$\pm$ 0.11} & 
\cellcolor{cyan!15}\textbf{38.06} {\tiny$\pm$ 0.21} & 
\cellcolor{cyan!15}\textbf{41.46} {\tiny$\pm$ 0.55} & 
\cellcolor{cyan!15}\textbf{41.63} {\tiny$\pm$ 0.21} \\
\bottomrule
\end{tabular}%
}
\end{table*}

\begin{table}[ht]
\centering
\caption{\textbf{Performance Results for NLP Task on MNLI, SST-2 and MARC.}}
\label{tab:fl_results_nlp}
\footnotesize
\begin{tabular}{lllcc}
\toprule
\textbf{Private} & \textbf{Public} & \textbf{Baseline} & \textbf{Dir($\beta$=0.3)} & \textbf{Dir($\beta$=0.5)} \\
\midrule

\multirow{5}{*}{MNLI \cite{mnli}} & \multirow{5}{*}{SNLI \cite{snli}} 
& FedAvg       & 35.67{\tiny$\pm$1.21} & 41.91{\tiny$\pm$3.98} \\
& & FedDF        & 36.65{\tiny$\pm$1.32} & 41.07{\tiny$\pm$5.88} \\
& & FedET        & 36.10{\tiny$\pm$3.34} & 36.50{\tiny$\pm$3.39} \\
& & \cellcolor{cyan!15}FedProj         & \cellcolor{cyan!15}\textbf{44.38}{\tiny$\pm$3.91} & \cellcolor{cyan!15}\textbf{45.13}{\tiny$\pm$3.10} \\
\midrule

\multirow{5}{*}{SST2 \cite{sst2}} & \multirow{5}{*}{Sent140 \cite{sent140}} 
& FedAvg       & 56.96{\tiny$\pm$1.36} & 55.08{\tiny$\pm$6.46} \\
& & FedDF        & 51.43{\tiny$\pm$2.19} & 54.45{\tiny$\pm$3.86} \\
& & FedET        & 54.96{\tiny$\pm$8.31} & 56.36{\tiny$\pm$9.42} \\
& & \cellcolor{cyan!15}FedProj         & \cellcolor{cyan!15}\textbf{64.80}{\tiny$\pm$5.1} & \cellcolor{cyan!15}\textbf{65.98}{\tiny$\pm$2.87} \\
\midrule

\multirow{5}{*}{MARC \cite{marc}} & \multirow{5}{*}{Yelp \cite{yelp}} 
& FedAvg       & 37.21{\tiny$\pm$2.85} & 40.86{\tiny$\pm$2.89} \\
& & FedDF        & 40.74{\tiny$\pm$2.91} & 38.40{\tiny$\pm$6.05} \\
& & FedET        & 37.02{\tiny$\pm$3.39} & 40.05{\tiny$\pm$2.94} \\
& & \cellcolor{cyan!15}FedProj         & \cellcolor{cyan!15}\textbf{45.15}{\tiny$\pm$1.59} & \cellcolor{cyan!15}\textbf{46.52}{\tiny$\pm$4.42} \\
\bottomrule
\end{tabular}

\end{table}
\noindent\textbf{Performance on NLP Task.} 
Table~\ref{tab:fl_results_nlp} presents results on three NLP private-public dataset pairs: MNLI--SNLI, SST-2--Sentiment140, and MARC--Yelp, under two Dirichlet non-IID settings ($\text{Dir}(\beta{=}0.3)$ and $\text{Dir}(\beta{=}0.5)$). Similar to CV results, on NLP experiments FedProj consistently outperforms existing baselines, demonstrating superior robustness across heterogeneous client scenarios. On MNLI~\cite{mnli}, paired with SNLI~\cite{snli}, FedProj delivers substantial gains. Under $\text{Dir}(\beta{=}0.3)$, it achieves 44.38 accuracy, surpassing FedDF (36.65) and FedET (36.10), with an improvement exceeding 7\% compared to FedAvg. These results expose the limitations of standard aggregation approaches under skewed data distributions. FedET underperforms FedAvg, revealing the downside of relying on uncertainty estimates from pretrained language models, which produce overconfident and poorly calibrated predictions~\cite{plm1, plm2, plm3}. For SST-2~\cite{sst2}--Sentiment140~\cite{sent140}, FedProj delivers state-of-the-art accuracy across both settings, improving by nearly 9\% over the strongest baseline. This proves FedProj's effectiveness in bridging domain gaps between curated sentiment labels and noisy social media text. On MARC~\cite{marc}--Yelp~\cite{yelp}, involving multilingual and domain-diverse reviews, FedProj maintains its dominance with 4--6\% gains across both non-IID scenarios. These results establish FedProj's effectiveness against both linguistic and domain shifts in realistic federated NLP applications.

\subsection{Ablation Studies}

\begin{figure}[t]
\centering

\begin{subfigure}[b]{0.48\textwidth}
    \centering
    \includegraphics[width=\linewidth]{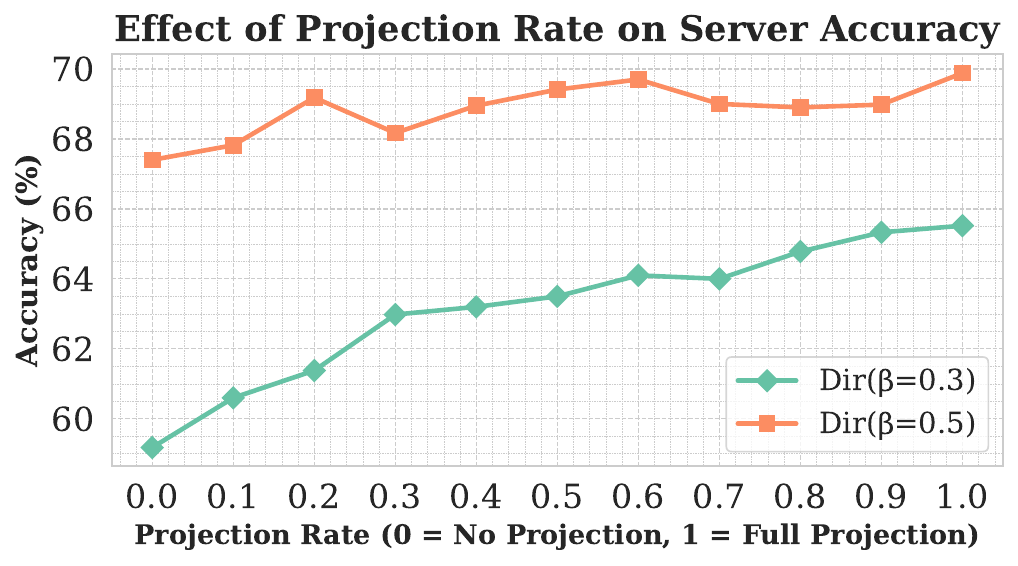}
    \caption{\textbf{Impact of Projection Dropout.}}
    \label{fig:projection_ablation}
\end{subfigure}
\hfill
\begin{subfigure}[b]{0.48\textwidth}
    \centering
    \footnotesize
    \begin{tabular}{@{}lccc@{}}
    \toprule
    \textbf{Dataset} & \textbf{WD} & \textbf{Dir($\beta$=0.3)} & \textbf{Dir($\beta$=0.5)} \\
    \midrule
    \multirow{4}{*}{CIFAR10}
    & 0.1 & 64.12{\tiny$\pm$1.42} & 66.57{\tiny$\pm$1.43} \\
    & 0.3 & 63.31{\tiny$\pm$0.29} & \textbf{69.88}{\tiny$\pm$0.03} \\
    & 0.5 & 63.07{\tiny$\pm$2.38} & 67.12{\tiny$\pm$1.85} \\
    & w/o & \textbf{65.52}{\tiny$\pm$0.86} & 67.23{\tiny$\pm$0.66} \\
    \midrule
    \multirow{4}{*}{CIFAR100}
    & 0.1 & 34.71{\tiny$\pm$0.16} & 37.67{\tiny$\pm$0.14} \\
    & 0.3 & 33.49{\tiny$\pm$0.24} & 36.99{\tiny$\pm$0.16} \\
    & 0.5 & 33.21{\tiny$\pm$0.31} & \textbf{38.06}{\tiny$\pm$0.21} \\
    & w/o & \textbf{35.27}{\tiny$\pm$0.11} & 36.98{\tiny$\pm$0.41} \\
    \midrule
    \multirow{4}{*}{CINIC10}
    & 0.1 & 39.85{\tiny$\pm$0.34} & 40.97{\tiny$\pm$0.22} \\
    & 0.3 & 40.11{\tiny$\pm$0.21} & \textbf{41.63}{\tiny$\pm$0.21} \\
    & 0.5 & 40.22{\tiny$\pm$0.36} & 41.25{\tiny$\pm$0.18} \\
    & w/o & \textbf{41.46}{\tiny$\pm$0.55} & 41.19{\tiny$\pm$0.27} \\
    \bottomrule
    \end{tabular}
    \caption{\textbf{Impact of Weight Divergence (WD).}}
    \label{tab:wd_ablation}
\end{subfigure}

\end{figure}

\textbf{Impact of Gradient Projection.}
We quantify client-side gradient projection's critical role on CIFAR-10, where projection is randomly omitted during client updates with varying rates. As shown in Fig.~\ref{fig:projection_ablation}, under high heterogeneity (\(\text{Dir}(\beta=0.3)\)), server accuracy increases sharply as projection rate increases, with full projection delivering superior performance. This proves that projection enforces alignment between local updates and the global objective, resulting into updates that directly minimize local loss while preserving the global knowledge. Even slight projection removal triggers substantial degradation as observed. Under lower data heterogeneity (\(\text{Dir}(\beta=0.5)\)), performance show stability at moderate projection levels but declines precipitously when projection is largely eliminated. These results establish projection as essential for global model alignment, especially in highly heterogeneous settings.

\textbf{Impact of Weight Divergence Regularization.} Table~\ref{tab:wd_ablation} presents the effect of L2 regularization on server-side performance for CIFAR-10, CIFAR-100, and CINIC-10. For CIFAR-10, the optional use of L2 regularization ($\lambda=0$) achieves the highest accuracy for $\beta=0.3$, while a moderate value ($\lambda=0.3$) performs best for $\beta=0.5$. In CIFAR-100, $\lambda=0.5$ yields the best result for $\beta=0.5$, whereas omitting L2 regularization is preferable for $\beta=0.3$. CINIC-10 shows minimal variation across values, indicating lower sensitivity to L2 regularization. Overall, the improvements from L2 regularization are marginal; the primary performance gains are attributed to knowledge fusion and the use of client-side gradient projection.

\section{Conclusion}
In this work, we propose FedProj, a federated learning framework designed to tackle catastrophic forgetting and enhance knowledge retention in non-IID settings. Specifically, FedProj consists of two key components: (1) a client-side gradient projection mechanism that preserves global knowledge by constraining local updates, preventing overfitting to client-specific objectives, and (2) a server-side knowledge distillation process that fuses local models through ensemble distillation on a small, auxiliary public dataset, effectively aligning their logits and feature representations. We conduct extensive experiments on benchmark datasets and demonstrate that FedProj outperforms state-of-the-art FL methods in terms of accuracy and stability on non-IID data, validating its effectiveness in preserving global knowledge while accommodating local adaptations. The limitations of our work are lack of real-world implementation and relying on public dataset which we will pursue in the future work.

\clearpage

\bibliographystyle{plainnat}
\bibliography{references}\
\clearpage
\appendix 

\section{Appendix}

\subsection{Detailed Derivation and Proof for Gradient Projection} \label{sec:appendix_proof}
We provide a detailed mathematical derivation of the gradient projection step used in FedProj.

\textbf{Lemma 1 (Preservation of Global Knowledge).} Given gradients $g_{\text{new}}$ from the local loss and $g_{\mathrm{glob}}$ from the memory-based global knowledge loss, the projection given by \eqref{eq:proj_grad_optimal} is the optimal solution to the constrained optimization problem stated in \eqref{eq:constraint_obj}.

\textbf{Proof.} Consider the optimization problem:
\begin{equation}
\label{eq:proj_problem_proof}
    \min_{g_{\text{proj}}} \quad \frac{1}{2}\|g_{\text{new}} - g_{\text{proj}}\|^2_2, \quad \text{s.t.} \quad \langle g_{\text{proj}}, g_{\mathrm{glob}}\rangle \geq 0.
\end{equation}

We introduce the Lagrangian:
\[
\mathcal{L}(g_{\text{proj}}, \lambda) = \frac{1}{2}\|g_{\text{new}} - g_{\text{proj}}\|^2_2 - \lambda \langle g_{\text{proj}}, g_{\mathrm{glob}}\rangle,
\]
with the Karush-Kuhn-Tucker (KKT) optimality conditions:
\begin{align}
    \frac{\partial\mathcal{L}}{\partial g_{\text{proj}}} &= g_{\text{proj}} - g_{\text{new}} - \lambda g_{\mathrm{glob}} = 0, \label{eq:kkt_grad}\\[4pt]
    \lambda &\geq 0, \\[4pt]
    \lambda \langle g_{\text{proj}}, g_{\mathrm{glob}}\rangle &= 0, \quad \langle g_{\text{proj}}, g_{\mathrm{glob}}\rangle \geq 0.
\end{align}
From the first KKT condition, we have:
\begin{equation}
\label{eq:kkt}
    g_{\text{proj}} = g_{\text{new}} + \lambda g_{\mathrm{glob}}.
\end{equation}

To find $\lambda$, we consider two cases:
\begin{itemize}
    \item \textbf{Case 1 ($\lambda = 0$):} If $\langle g_{\text{new}},\,g_{\mathrm{glob}}\rangle \geq 0$, then $g_{\text{proj}} = g_{\text{new}}$ directly satisfies the constraint, making it the optimal solution.
    
    \item \textbf{Case 2:} If $\langle g_{\text{new}}, g_{\mathrm{glob}}\rangle < 0$, the constraint becomes active:
\[
    \langle g_{\text{proj}},\,g_{\mathrm{glob}}\rangle = 0.
\]
\end{itemize}
Substituting \eqref{eq:kkt}, we obtain:
\[
    \langle g_{\text{new}} + \lambda g_{\mathrm{glob}}, g_{\mathrm{glob}}\rangle = 0 \quad\Rightarrow\quad \lambda = -\frac{\langle g_{\text{new}}, g_{\mathrm{glob}}\rangle}{\|g_{\mathrm{glob}}\|^2 + \epsilon}.
\]
Hence, the optimal projected gradient is:
\begin{equation}
\label{eq:proj_grad_optimal}
    g_{\text{proj}} = g_{\text{new}} - \frac{\langle g_{\text{new}},\,g_{\mathrm{glob}}\rangle}{\|g_{\mathrm{glob}}\|^2 + \epsilon}\,g_{\mathrm{glob}},
\end{equation}
matching Eq.~\eqref{eq:proj_grad_optimal}. Thus, the derived gradient projection update guarantees that local model updates do not negatively affect global knowledge retention.

\newpage

\subsection{Full Algorithm Description of FedProj}
The full algorithm description of FedProj is presented in Algorithm \ref{alg:FedProj}.

\begin{algorithm}[t!]
\caption{\textbf{FedProj: Federated Learning with Gradient Projection and Distillation}}
\label{alg:FedProj}
\small
\begin{algorithmic}[1]
\REQUIRE Number of rounds $T$, local learning rate $\eta_{\mathrm{local}}$, distillation rate $\eta_{\mathrm{distill}}$, local epochs $E$, distillation epochs $E_d$, public dataset $\mathcal{D}_{\mathrm{pub}}$, memory-based gradient constraint.
\STATE Initialize server model parameters $\btheta_g^{(0)}$.
\FOR{$t=0,\dots,T-1$}
    \STATE Sample a subset $\mathcal{S}_t$ of clients
    \FORALL{$k\in \mathcal{S}_t$ \textbf{in parallel}}
        \STATE $\btheta_k \gets \btheta_g^{(t)}$ \textit{Initialize local model}
        \FOR{epoch $e=1,\dots,E$}
            \FOR{mini-batch $(\mathbf{x},\mathbf{y})\subseteq \mathcal{D}_k$}
                \STATE $g_{\text{new}} \,\gets\, \nabla_{\btheta_k}\ell\bigl(f(\mathbf{x};\btheta_k),\,\mathbf{y}\bigr)$
               \STATE $g_{\mathrm{glob}} \,\gets\, \nabla_{\btheta_k} \bigl[\tfrac{1}{|\mathcal{D}_{\mathrm{pub}}|} \hspace{-0.2cm}\sum\limits_{(\mathbf{x}_m,\mathbf{y}_m)\in \mathcal{D}_{\mathrm{pub}}} \hspace{-0.7cm}\ell\bigl(f(\mathbf{x}_m;\btheta_k),\mathbf{y}_m\bigr)\bigr]$
                \STATE Project as in \eqref{eq:proj_grad_optimal}: $g_{\text{proj}} \gets g_{\text{new}} - \frac{\langle g_{\text{new}},\,g_{\mathrm{glob}}\rangle}{\|g_{\mathrm{glob}}\|^2 + \epsilon}\,g_{\mathrm{glob}}$
                \STATE $\btheta_k \;\gets\; \btheta_k \;-\; \eta_{\mathrm{local}}\; g_{\text{proj}}$
            \ENDFOR
        \ENDFOR
        \STATE $\btheta_k^{(t+1)} \gets \btheta_k$
    \ENDFOR
    \STATE \emph{// \underline{Server aggregation \& distillation}}
    \STATE $\btheta_g^{\mathrm{FedAvg}} \gets \sum_{k\in \mathcal{S}_t}\!\Bigl(\frac{|\mathcal{D}_k|}{\sum_{j\in \mathcal{S}_t}|\mathcal{D}_j|}\Bigr)\btheta_k^{(t+1)}$
    \STATE Initialize $\btheta_g \gets \btheta_g^{\mathrm{FedAvg}}$
    \FOR{epoch $e_d=1,\dots,E_d$}
        \FOR{mini-batch $\mathbf{X}\subseteq \mathcal{D}_{\mathrm{pub}}$}
            \STATE Compute teacher logits: 
            \[ \hspace{-2cm}
               \mathcal{Z}_{\mathrm{teacher}} = \frac{1}{|\mathcal{S}_t|} \sum_{k\in \mathcal{S}_t} f(\mathbf{X};\,\btheta_k^{(t+1)})
            \]
            \STATE $\mathcal{Z}_{\mathrm{student}} = f(\mathbf{X};\btheta_g)$
            \STATE $\mathcal{L}_{\mathrm{KD}} = T^2\cdot \mathrm{KL}\bigl(\sigma(\tfrac{\mathcal{Z}_{\mathrm{student}}}{T}),\,\sigma(\tfrac{\mathcal{Z}_{\mathrm{teacher}}}{T})\bigr)$  
            \STATE $\mathcal{L}_{\mathrm{div}} = \alpha\,\|\btheta_g - \btheta_g^{(t)}\|_2^2$  
            \STATE $\btheta_g \;\gets\; \btheta_g \;-\;\eta_{\mathrm{distill}}\,\nabla_{\btheta_g}\bigl[\mathcal{L}_{\mathrm{KD}} + \mathcal{L}_{\mathrm{div}}\bigr]$

        \ENDFOR
    \ENDFOR
    \STATE $\btheta_g^{(t+1)} \gets \btheta_g$
\ENDFOR
\end{algorithmic}
\end{algorithm}







\end{document}